\documentclass[sigconf,nonacm]{acmart}

\DeclareUnicodeCharacter{2212}{-}

\renewcommand\footnotetextcopyrightpermission[1]{} 
\acmConference[]{ }{ }{ }   
\acmBooktitle{}
\acmPrice{}
\acmISBN{}
\acmDOI{}
\copyrightyear{}

\usepackage{float}
\usepackage{caption}
\usepackage{subcaption}
\usepackage[most]{tcolorbox}
\tcbset{
  colback=gray!5,
  colframe=black!75,
  arc=4pt,
  boxrule=0.7pt,
  left=6pt,right=6pt,top=6pt,bottom=6pt
}

\newtcolorbox{promptbox}[1][]{
  breakable,
  enhanced,
  fonttitle=\bfseries,
  colbacktitle=black!80,
  coltitle=white,
  title=#1
}

\tcbset{
  baselinebox/.style={
    enhanced,
    breakable,
    colback=gray!10,
    colframe=black!75
  }
}

\begin{document}

\title{OntoMetric: An Ontology-Driven LLM-Assisted Framework for Automated ESG Metric Knowledge Graph Generation}

\author{Mingqin Yu}
\authornote{Corresponding author.}
\email{mingqin.yu@unsw.edu.au}
\affiliation{%
  \institution{School of Computer Science and Engineering, University of New South Wales}
  \city{Sydney}
  \country{Australia}
}

\author{Fethi Rabhi}
\email{f.rabhi@unsw.edu.au}
\affiliation{%
  \institution{School of Computer Science and Engineering, University of New South Wales}
  \city{Sydney}
  \country{Australia}
}

\author{Boming Xia}
\email{boming.xia@adelaide.edu.au}
\affiliation{%
  \institution{Faculty of Sciences, Engineering and Technology, The University of Adelaide}
  \city{Adelaide}
  \country{Australia}
}

\author{Zhengyi Yang}
\email{zhengyi.yang@unsw.edu.au}
\affiliation{%
  \institution{School of Computer Science and Engineering, University of New South Wales}
  \city{Sydney}
  \country{Australia}
}

\author{Felix Tan}
\email{f.tan@unsw.edu.au}
\affiliation{%
  \institution{School of Information Systems and Technology Management, University of New South Wales}
  \city{Sydney}
  \country{Australia}
}

\author{Qinghua Lu}
\email{qinghua.lu@data61.csiro.au}
\affiliation{%
  \institution{CSIRO's Data61}
  \city{Sydney}
  \country{Australia}
}

\begin{abstract}
Environmental, Social, and Governance (ESG) metric knowledge is inherently
structured, connecting industries, reporting frameworks, metric categories,
metrics, and calculation models through compositional dependencies, yet in
practice this structure remains embedded implicitly in regulatory documents such
as SASB, TCFD, and IFRS~S2 and rarely exists as an explicit, governed, or
machine-actionable artefact. Existing ESG ontologies define formal schemas but do
not address scalable population and governance from authoritative regulatory
sources, while unconstrained large language model (LLM) extraction frequently
produces semantically incorrect entities, hallucinated relationships, and
structurally invalid graphs. OntoMetric is an ontology-guided framework for the
automated construction and governance of ESG metric knowledge graphs from
regulatory documents that operationalises the ESG Metric Knowledge Graph
(ESGMKG) ontology as a first-class constraint embedded directly into the
extraction and population process. The framework integrates structure-aware
segmentation, ontology-constrained LLM extraction enriched with semantic fields
and deterministic identifiers, and two-phase validation combining semantic type
verification with rule-based schema checking, while preserving segment-level and
page-level provenance to ensure traceability to regulatory source text.
Evaluation on five ESG regulatory standards shows that ontology-guided
extraction achieves 65--90\% semantic accuracy and over 80\% schema compliance,
compared with 3--10\% for unconstrained baseline extraction, and yields stable
cost efficiency with a cost per validated entity of \$0.01--\$0.02 and a
48$\times$ efficiency improvement over baseline.
\end{abstract}


\maketitle

\section{Introduction}

Environmental, Social, and Governance (ESG) metrics constitute a structured body of domain knowledge that specifies what must be measured, how values are computed, which units apply, and how individual indicators depend on one another. Beyond simple numerical values, ESG metric knowledge includes formal definitions, scope and boundary conditions, disaggregation rules, calculation models, and compositional dependencies between metrics and their input variables. Collectively, these elements form an implicit \emph{ESG metric knowledge graph} that connects industries, reporting contexts, metric categories, metrics, and computational models into a coherent semantic structure.

In practice, however, this ESG metric knowledge graph does not exist as an explicit, governed, or machine-actionable artefact. Instead, metric definitions, calculation logic, and inter-metric dependencies are embedded implicitly in regulatory documents and reporting artefacts. Where ESG ontologies have been proposed, they typically provide only formal schemas or high-level concept hierarchies, without populated instances or traceable links to authoritative metric definitions. As a result, the construction and maintenance of ESG metric knowledge graphs remains a largely manual, expert-driven process.

Several efforts have sought to formalise ESG knowledge using ontologies and semantic models. Notably, we previously proposed the ESG Metric Knowledge Graph (ESGMKG), an ontology-driven architecture that defines the core entity types, relationships, and compositional structures required to represent ESG metric knowledge, including industries, reporting frameworks, metric categories, metrics, and calculation models \cite{yu2024_esgmk}. However, ESGMKG and related ontologies focus primarily on schema definition rather than scalable population and governance. They do not address how ESG metric knowledge can be constructed automatically from regulatory sources or how provenance can be preserved at scale.

In real-world settings, ESG metric knowledge is derived from multiple sources, among which regulatory and quasi-regulatory reporting frameworks such as the Sustainability Accounting Standards Board (SASB) \cite{sasb}, the Task Force on Climate-related Financial Disclosures (TCFD) \cite{tcfd}, and the International Sustainability Standards Board’s IFRS~S2 Climate-related Disclosures \cite{ifrs_s2} remain the most authoritative and widely adopted references for defining what should be disclosed and how metrics should be interpreted. However, the metric knowledge they provide is rarely expressed in an explicit, machine-actionable form. Definitions, scope conditions, units, and dependencies are distributed across narrative text, tables, and cross-referenced sections, and the level of computational specificity varies substantially across frameworks.

Taken together, these observations expose a fundamental automation and governance gap in the construction of ESG metric knowledge graphs. Although ESG metric knowledge is inherently structured and graph-like, and although formal ontologies exist to model its schema, there is currently no scalable, reliable, and auditable mechanism for populating such graphs from authoritative regulatory sources. Prior attempts to automate this process using keyword matching, handcrafted rules, or traditional named-entity recognition pipelines lack the semantic coverage required to recover compositional metric definitions and dependencies embedded in regulatory text \cite{adnan2019_ie_limitations,al2020_ner_kg_survey}. Large language models (LLMs) offer strong natural-language understanding capabilities \cite{bommasani2021_foundation_models,pan2024_llm_kg_roadmap}, but unconstrained LLM-based extraction frequently produces semantically incorrect entities and hallucinated relationships, making the resulting graphs unsuitable for regulatory and audit contexts \cite{ji2023_hallucinations}.

These limitations indicate that the core challenge is not simply extracting entities from regulatory documents, but constructing and governing ESG metric knowledge graphs with guarantees of semantic fidelity, structural validity, and traceable provenance. Addressing this challenge requires a framework that treats the ontology not merely as a schema for post-hoc validation, but as a first-class constraint embedded directly into the extraction and population process.

To address this automation and governance gap, we present \textbf{OntoMetric}, an ontology-guided framework for the automated construction and governance of ESG metric knowledge graphs from regulatory documents. OntoMetric operationalises an ontology-first automation strategy in which the ESG Metric Knowledge Graph (ESGMKG) ontology is treated as a first-class constraint embedded directly into the extraction and population process, rather than as a schema applied only for post-hoc validation. OntoMetric integrates structure-aware segmentation, ontology-constrained large language model (LLM) extraction enriched with semantic fields, and two-phase validation combining semantic type verification with rule-based schema checking. Throughout this process, OntoMetric preserves segment-level and page-level provenance links to authoritative regulatory sources, ensuring auditability and traceability.

\paragraph{Contributions.}
This paper addresses the problem of scalable, auditable construction of ESG metric knowledge graphs from unstructured regulatory sources. We make the following three contributions:

\begin{itemize}
  \item \textbf{Ontology-First ESG Metric Knowledge Graph Construction.}  
  We propose OntoMetric, an ontology-guided framework that treats the ESG Metric Knowledge Graph (ESGMKG) ontology as a first-class constraint embedded directly into the extraction and population process, enabling the automated construction and governance of ESG metric knowledge graphs while preserving compositional dependencies between metrics, models, and input variables.

  \item \textbf{Two-Phase Validation and Provenance-Preserving Governance.}  
  We introduce a two-phase validation architecture that combines semantic type verification with rule-based schema checking to enforce semantic fidelity, structural validity, and traceable provenance across entities, properties, and relationships.

  \item \textbf{Empirical Evaluation on Real-World ESG Standards.}  
  We conduct a systematic evaluation of OntoMetric on five ESG regulatory documents (SASB Commercial Banks, SASB Semiconductors, TCFD, IFRS
\end{itemize}

\section{Related Work}
\subsection{ESG Metric Ontologies and Knowledge Representation}

Efforts to formalise Environmental, Social, and Governance (ESG) knowledge using ontologies and semantic models have progressed from general sustainability vocabularies to more specialised representations targeting corporate disclosure requirements \cite{konys2018_sustainability_ontology, zhou2023_ontosustain}. These ontologies aim to standardise terminology and represent reporting relationships, but they typically operate at the level of conceptual taxonomies and entity hierarchies, without explicitly modelling the internal semantic structure of ESG metrics, including formal definitions, calculation logic, scope conditions, and compositional dependencies between metrics and their input variables.

This limitation is critical for ESG metric knowledge graphs, where metrics are interdependent semantic objects whose meaning is defined through relationships to reporting contexts, calculation models, and required input variables. Representing such knowledge therefore requires a schema that can capture how metrics are derived and how they depend on one another, rather than only high-level sustainability concepts.

To address this representational gap, we previously proposed the ESG Metric Knowledge Graph (ESGMKG), an ontology-driven architecture that defines the core entity types and relationships required to represent ESG metric knowledge \cite{yu2024_esgmk}. ESGMKG formalises how calculated indicators are derived from input variables and how disclosure requirements relate across reporting standards, providing a principled schema for representing ESG metric semantics in a graph-structured form.

However, ESGMKG and related ESG metric ontologies focus primarily on schema definition rather than scalable population and governance. They do not specify how ESG metric knowledge can be constructed automatically from regulatory documents, how provenance links to authoritative sources can be preserved at scale, or how evolving standards can be reconciled with existing knowledge graph instances. As a result, although such ontologies provide the conceptual foundation for ESG metric knowledge graphs, they do not resolve the core bottleneck of constructing and maintaining these graphs from unstructured regulatory text.

Moreover, most ESG ontologies represent entities as identifiers with basic type information, without the richer semantic descriptions needed to support downstream reasoning and integration with modern AI systems \cite{hogan2021_kg_survey}. This further underscores the need for frameworks that not only define the schema of ESG metric knowledge, but also enable its semantic enrichment, governed population, and provenance-preserving construction from regulatory sources.

\subsection{Construction and Validation of ESG Metric Knowledge Graphs}

Knowledge graph construction methods range from manual curation to automated extraction pipelines. Traditional automated approaches rely on named entity recognition and relation extraction trained on domain-specific corpora \cite{al2020_ner_kg_survey}, but they struggle with regulatory ESG documents in which metric definitions span tables, multi-paragraph narratives, and compositional formulas \cite{adnan2019_ie_limitations}. As a result, existing ESG metric ontologies such as ESGMKG rely on manual population, limiting their scalability across evolving reporting standards \cite{yu2024_esgmk}.

Large language models (LLMs) offer strong zero-shot extraction capabilities that substantially reduce manual effort \cite{carta2023_iterative_zeroshot}, and they have recently been applied to knowledge graph construction \cite{pan2024_llm_kg_roadmap}. However, unconstrained LLM-based extraction frequently produces semantically incorrect entity types, invalid relationships, and hallucinated content \cite{ji2023_hallucinations}, making the resulting graphs unsuitable for regulatory and audit contexts. Without ontology-level constraints, such outputs cannot be trusted as governed representations of ESG metric knowledge.

Ontology-guided extraction has been explored in domains such as biomedicine \cite{lee_biobert} and law \cite{chalkidis2020_legalbert}, where predefined schemas are used to constrain entity typing and relation extraction. These approaches improve structural consistency but typically stop at extraction, with limited support for post-extraction validation, provenance preservation, or governance. As a result, they do not satisfy the quality and auditability requirements of ESG metric knowledge graphs.

Validation methods for knowledge graphs generally fall into two categories: schema-level constraints (e.g., required fields, valid predicates, cardinality rules) and semantic-level verification (e.g., type correctness, factual accuracy) \cite{paulheim2017_kg_refinement, zaveri2016_quality_assessment}. Most extraction pipelines enforce only schema constraints, leaving semantic validation to manual review—an approach that does not scale for regulatory contexts requiring traceability and auditability. This exposes a critical gap: while techniques exist for extracting triples from text and checking basic schema compliance, there is no scalable, integrated framework for constructing ESG metric knowledge graphs with guarantees of semantic fidelity, structural validity, and provenance preservation.

\section{The OntoMetric Framework Design}
\label{sec:framework-design}

\subsection{Overview of the OntoMetric Framework}

\begin{figure*}[htbp]
  \centering
  \includegraphics[width=0.9\linewidth]{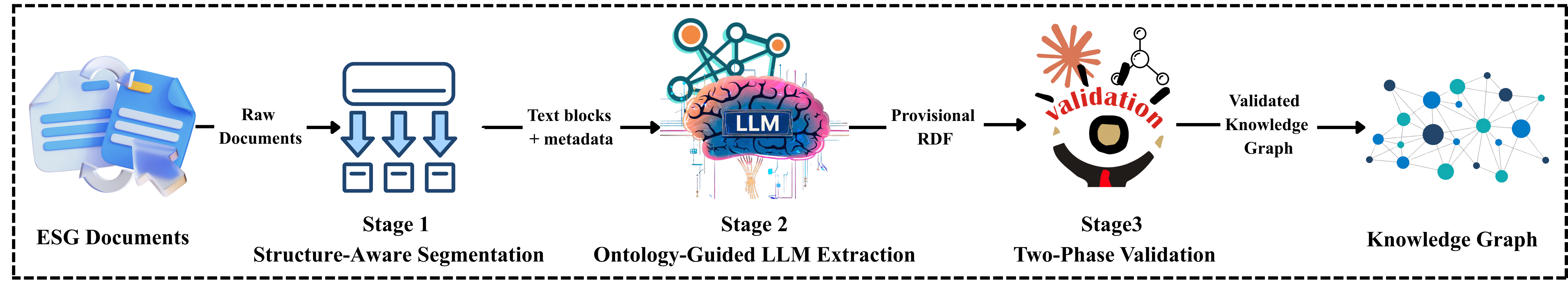}
  \caption{OntoMetric Framework}
  \Description{A diagram showing the three-stage OntoMetric workflow with inputs, intermediate artifacts, and outputs.}
  \label{fig:framework-overview}
\end{figure*}

OntoMetric is an ontology-first, three-stage framework for the automated construction and governance of ESG metric knowledge graphs from unstructured regulatory documents. Its primary objective is to transform narrative ESG standards into validated, provenance-preserving knowledge graph instances that comply with the ESG Metric Knowledge Graph (ESGMKG) ontology. As shown in Figure~\ref{fig:framework-overview}, OntoMetric integrates: (1)~structure-aware document segmentation to isolate coherent semantic units, (2)~ontology-constrained large language model (LLM) extraction to instantiate ESGMKG entities and relationships, and (3)~two-phase validation combining semantic type verification with rule-based schema checking.

Each stage addresses a distinct requirement derived from the automation and governance challenges identified in Sections~1 and~2. Stage~1 segments lengthy regulatory PDFs into context-preserving units while retaining structural and provenance metadata needed for auditability. Stage~2 applies ESGMKG as a first-class constraint on extraction, guiding the LLM to produce schema-aligned entities, relationships, and compositional dependencies from each segment. Stage~3 enforces knowledge graph quality through a two-phase validation process: semantic verification filters type and role inconsistencies at the entity level, followed by rule-based validation that checks compliance with ESGMKG structural constraints.

This modular, ontology-guided design yields interpretable intermediate artefacts, consistent behaviour across heterogeneous regulatory documents, and explicit provenance links to authoritative sources. The resulting knowledge graphs are ESGMKG-compliant, semantically enriched with labels, contextual descriptions, and structured properties, and suitable for downstream applications such as automated compliance analysis, cross-framework metric alignment, and AI-assisted ESG reasoning.

\subsection{Structure-Aware Segmentation}

OntoMetric begins by partitioning each regulatory document into coherent, context-aligned segments that serve as the semantic units for downstream ontology-constrained extraction. ESG standards such as SASB and IFRS~S2 are organised hierarchically by disclosure topics (e.g., ``Data Security'', ``Greenhouse Gas Emissions''), with each section embedding metric definitions, calculation methodologies, reporting scope, and narrative guidance. Preserving these natural document boundaries is essential for maintaining the semantic integrity of metric knowledge and for ensuring traceable links between extracted entities and their authoritative sources.

OntoMetric performs structure-aware segmentation using table-of-contents (TOC) page detection and section title extraction to identify meaningful segmentation boundaries. This TOC-aligned strategy avoids fixed-size chunking, which frequently fragments coherent regulatory content and disrupts cross-referenced definitions in legal and technical documents. By aligning segments with the logical structure of the source document, OntoMetric ensures that each segment corresponds to a complete conceptual unit, such as a metric definition, a compositional formula, or a category-level description.

For each detected section, OntoMetric extracts all narrative text and tables within the associated page range. Tables spanning multiple pages are merged, and light layout cleanup is applied to remove artefacts introduced during PDF parsing. Each resulting segment preserves structural and provenance metadata, including section titles, page spans, and section identifiers. This metadata is retained throughout subsequent stages to support auditability, traceability, and the construction of provenance-preserving knowledge graph instances.

Prior work has shown that context-preserving segmentation improves information extraction accuracy by maintaining semantic coherence and reducing fragmentation. OntoMetric adopts this principle not as a generic pre-processing step, but as a governance requirement for regulatory-grade ESG metric knowledge graphs, ensuring that downstream ontology-constrained extraction operates over semantically complete and provenance-linked document units.

\subsection{Ontology-Constrained LLM Extraction}

Stage~2 transforms each context-preserving segment produced in Stage~1 into structured ESG metric knowledge by instantiating ESGMKG-compliant entities and relationships through ontology-constrained LLM extraction. Rather than treating the language model as an unconstrained extractor, OntoMetric uses the ESG Metric Knowledge Graph (ESGMKG) ontology as a first-class semantic contract that governs what entities may be produced, how they may be related, and which structural constraints must be satisfied.

\begin{figure}[t]
  \centering
  \fbox{%
    \includegraphics[width=0.95\columnwidth]{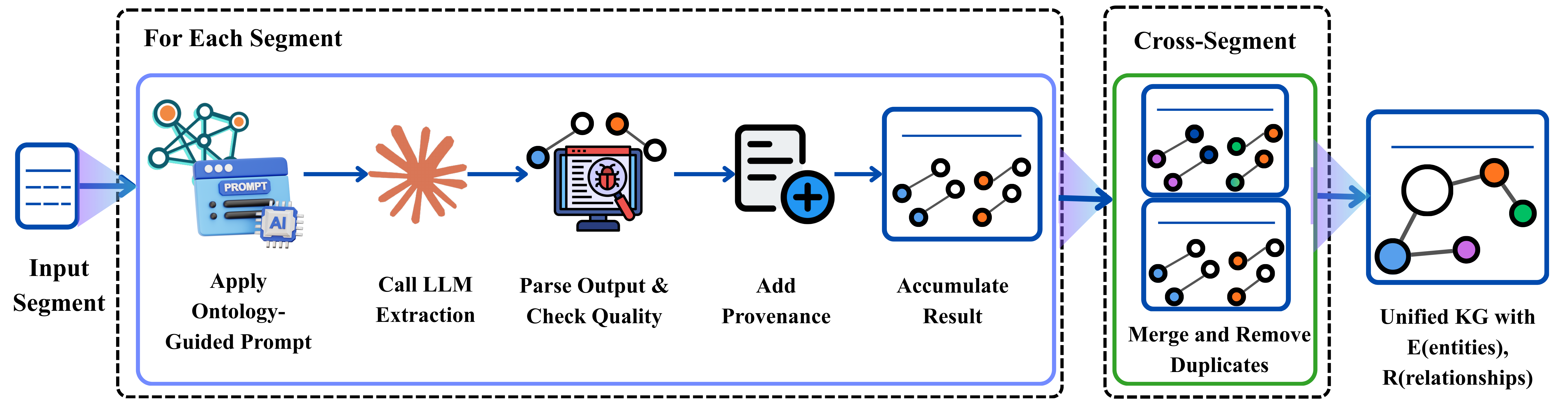}%
  }
  \caption{Stage~2: Ontology-constrained LLM extraction.}
  \label{fig:stage2-extraction}
\end{figure}

As illustrated in Figure~\ref{fig:stage2-extraction}, extraction proceeds in two phases: segment-level instantiation followed by cross-segment consolidation. At the segment level, each input segment is processed independently to generate candidate entities, relationships, and provenance annotations. These intermediate results are then merged across segments to form a unified ESG metric knowledge graph instance.

The ESGMKG ontology \cite{yu2024_esgmk} defines five core entity types: \emph{Industry}, \emph{ReportingFramework}, \emph{Category}, \emph{Metric}, and \emph{Model}, together with relationship types capturing regulatory hierarchies and metric computation dependencies. OntoMetric extends this schema with semantically enriched fields, including human-readable labels, contextual descriptions, and structured domain properties, to support downstream reasoning and interpretation. The resulting ontology structure defines both the permissible shape of ESG metric knowledge graphs and the semantic content that must be extracted for each entity instance.

For each segment, OntoMetric applies an ontology-guided prompt that combines the segment content with ESGMKG schema definitions, allowed predicates, and cardinality rules. The language model is instructed to generate structured outputs conforming to a predefined JSON schema, producing entities annotated with semantic fields and relationships grounded in the ontology. Each extracted entity is augmented with provenance metadata linking it to its source segment and page range \cite{prov_o_specification}. To support monitoring and governance, the system performs lightweight extraction checks at this stage, including type validity and field completeness, before passing candidates to downstream validation.

The ontology-guided prompt template comprises nine components (Table~\ref{tab:prompt-components}) encoding ESGMKG constraints, extraction workflow steps, and output structure. These components operationalise the ontology as an explicit interface between unstructured regulatory text and structured ESG metric knowledge, constraining the LLM to produce schema-aligned outputs rather than unconstrained triples.

\begin{table}[t]
\centering
\small
\caption{Components of the ontology-guided extraction prompt.}
\label{tab:prompt-components}
\begin{tabular}{@{}p{0.30\columnwidth} p{0.66\columnwidth}@{}}
\toprule
\textbf{Component} & \textbf{Description} \\
\midrule
System context & Establishes the LLM's role as an ESG metric knowledge graph extractor and defines the extraction objective. \\

Ontology schema & Specifies the five ESGMKG entity types (\emph{Industry}, \emph{ReportingFramework}, \emph{Category}, \emph{Metric}, \emph{Model}) and five relationship predicates (\emph{ReportUsing}, \emph{Include}, \emph{ConsistOf}, \emph{IsCalculatedBy}, \emph{RequiresInputFrom}). \\

Entity properties & Defines the required semantic fields for each entity type (e.g., a \emph{Metric} requires \texttt{measurement\_type}, \texttt{metric\_type}, \texttt{unit}, \texttt{code}, and \texttt{description}). \\

Relationship rules & Specifies permitted connections between entity types together with cardinality constraints (e.g., \emph{Category} $\rightarrow$ \emph{Metric} via \emph{ConsistOf}). \\

ID generation rules & Defines deterministic naming conventions for entity identifiers (e.g., \texttt{metric\_\{docid\}\_\{page\}\_\{nn\}}). \\

Extraction workflow & Describes the four-step process: (1) detect categories, (2) identify metrics, (3) identify models, and (4) validate the extracted output. \\

Output schema & Specifies the required JSON structure with mandatory fields and typing constraints for \texttt{meta}, \texttt{entities}, and \texttt{relationships}. \\

Few-shot examples & Provides concrete extraction examples distinguishing \emph{DirectMetric} from \emph{CalculatedMetric}. \\

Segment content & Appends the Stage~1 document segment text to be processed. \\

\bottomrule
\end{tabular}
\end{table}

After all segments have been processed, OntoMetric applies a three-pass consolidation procedure to produce a unified knowledge graph $G = (E, R)$. Identifier resolution normalises entity identifiers across segments, entity deduplication merges duplicate entities, and relationship deduplication removes redundant triples \cite{paulheim2017_kg_refinement}. This consolidation step reconciles segment-level extractions into a coherent, document-level ESG metric knowledge graph instance.

This design realises a reusable pattern for ontology-constrained LLM extraction. Segment-level processing ensures that each extraction operates over a semantically coherent unit; ontology constraints embedded directly in the prompt guide the LLM toward structurally valid outputs without relying on post-hoc filtering \cite{pan2024_llm_kg_roadmap}; and provenance annotation at extraction time preserves traceable links to authoritative regulatory text. Together, these mechanisms provide the semantic discipline and governance guarantees required for constructing ESG metric knowledge graphs suitable for regulatory and audit contexts.

\subsection{Two-Phase Validation and Governance}

\begin{figure}[t]
  \centering
  \includegraphics[width=\linewidth]{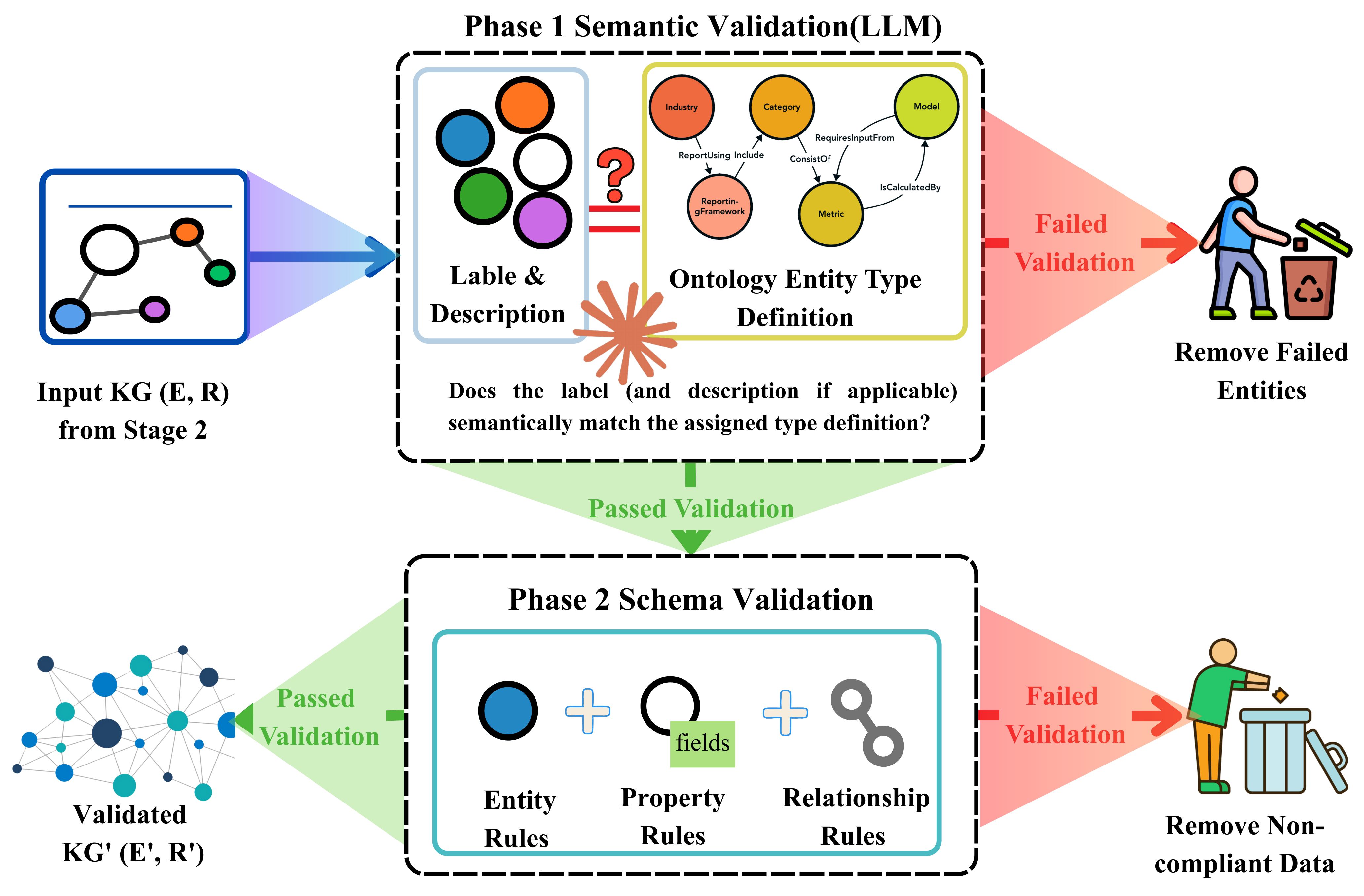}
  \caption{Stage~3: Two-phase validation and governance. Phase~1 applies LLM-based
  semantic verification to ensure correct ESGMKG type assignments. Phase~2 enforces
  ESGMKG schema rules (VR001--VR006), removing invalid entities and relationships to
  produce the validated graph $G' = (E', R')$.}
  \label{fig:stage3-validation}
\end{figure}

After consolidation, Stage~3 performs two-phase validation
(Figure~\ref{fig:stage3-validation}) that combines semantic verification with
rule-based schema checking. While ontology-constrained extraction substantially
improves output consistency, extraction alone is insufficient for regulatory and
audit contexts, where incorrectly typed entities, invalid dependencies, or missing
required fields can compromise compliance and traceability. OntoMetric therefore
introduces a two-phase validation architecture that embeds explicit governance and
quality control directly into the knowledge graph construction pipeline.

\textbf{Phase~1: Semantic Verification.}
In Phase~1, an LLM verifies whether each extracted entity is semantically consistent
with its assigned ESGMKG type based on its label, description, and domain
properties. For example, candidate \emph{Metric} entities are checked to ensure
that they correspond to measurable disclosure indicators, while \emph{Model}
entities are verified to represent legitimate calculation or aggregation
procedures. Entities belonging to unknown or implausible types are rejected
immediately to reduce unnecessary API usage. Entities failing semantic
verification are removed together with all dependent relationships, preventing
semantically invalid objects from propagating into the final knowledge graph.

\textbf{Phase~2: Schema Validation.}
In Phase~2, OntoMetric applies a suite of rule-based validators that enforce
ESGMKG structural constraints across entities, properties, and relationships.
These validators guarantee that the resulting graph instance conforms strictly
to the ontology schema.

\begin{table}[t]
\centering
\small
\caption{Rule-based validators grouped by validation type.}
\label{tab:validation-groups}
\begin{tabular}{@{}l p{0.28\linewidth} p{0.55\linewidth}@{}}
\toprule
\textbf{ID} & \textbf{Name} & \textbf{Description} \\
\midrule
\multicolumn{3}{@{}l}{\textbf{Entity Validation}} \\
VR001 & ID Uniqueness & Entity identifiers must be globally unique. \\
VR002 & Required Fields & Entities must contain all type-specific mandatory fields. \\
\midrule
\multicolumn{3}{@{}l}{\textbf{Property Validation}} \\
VR003 & Metric Values & Metric \texttt{code} and \texttt{unit} must be non-empty. \\
VR004 & Model Inputs & Models must reference at least one valid input variable. \\
\midrule
\multicolumn{3}{@{}l}{\textbf{Relationship Validation}} \\
VR005 & Predicate Validity & Only ESGMKG predicates are permitted. \\
VR006 & CM--Model Link & Each \emph{CalculatedMetric} must link to exactly one \emph{Model}. \\
\bottomrule
\end{tabular}
\end{table}

The six validators in Table~\ref{tab:validation-groups} enforce ontology
constraints at three complementary levels: entity-level checks (unique
identifiers, required fields), property-level checks (valid metric values,
non-empty model inputs), and relationship-level checks (approved predicates,
one-to-one links between \emph{CalculatedMetric} and \emph{Model}). Violations
trigger systematic removal of invalid entities and relationships together with
detailed validation logs.

This two-phase architecture separates concerns. Semantic verification captures
conceptual errors that schema rules cannot detect, such as plausible-sounding but
incorrect type assignments. Schema validation enforces formal structural
constraints that LLMs may overlook. Their combination yields a governed
construction pipeline that is both expressive enough to capture semantic nuance
and precise enough to guarantee structural correctness.

\paragraph{Validation Metrics and Governance Instrumentation.}
Beyond filtering invalid content, Stage~3 also instruments the construction
process with explicit quality and efficiency metrics that support auditability,
comparative evaluation, and reproducibility. These metrics are computed after
validation completes and are treated as first-class outputs of OntoMetric.

\noindent\textbf{Semantic Accuracy.}  
Measures the proportion of entities whose assigned ESGMKG types are semantically
correct, as determined by Phase~1 verification:
\[
\text{Semantic Accuracy} =
\frac{\text{Semantically Correct Entities}}{\text{Total Extracted Entities}} \times 100\%.
\]

\noindent\textbf{Schema Compliance.}  
Measures structural validity as the average compliance rate across the six
rule-based validators (VR001--VR006). Each rule score is computed as the
percentage of items passing that rule:
\[
\text{Schema Compliance} =
\frac{1}{6} \sum_{i=1}^{6}
\frac{\text{Items Passing VR}_i}{\text{Total Items Evaluated for VR}_i}
\times 100\%.
\]

\noindent\textbf{Relationship Retention.}  
Indicates the proportion of extracted relationships that survive validation,
reflecting the strictness of the governance constraints:
\[
\text{Relationship Retention} =
\frac{\text{Validated Relationships}}{\text{Extracted Relationships}} \times 100\%.
\]

\noindent\textbf{Cost per Entity.}  
Measures the API cost incurred in Stage~2 extraction and Stage~3 semantic
verification to produce one validated entity:
\[
\text{Cost per Entity} =
\frac{\text{Stage~2 Cost} + \text{Stage~3 Cost}}{\text{Validated Entities}}
\quad (\text{USD}).
\]

\noindent\textbf{Cost Waste Ratio.}  
Captures the proportion of extraction effort spent on entities that are
subsequently filtered during validation:
\[
\text{Cost Waste Ratio} =
\frac{\text{Filtered Entities}}{\text{Total Extracted Entities}} \times 100\%.
\]
Lower values indicate more efficient extraction, with fewer wasted API calls on
semantically or structurally invalid entities.

These metrics transform validation from a purely technical filtering
step into a governance and accountability mechanism. In addition to producing a
validated knowledge graph $G' = (E', R')$, OntoMetric outputs a detailed validation
log and a quantitative quality profile, enabling reproducible construction,
cross-document comparison, and regulatory audit support.

\section{Evaluation}
\subsection{Research Questions}

The evaluation addresses three research questions, each targeting a single,
distinct quality dimension of the OntoMetric framework and operationalised using
a non-overlapping set of evaluation metrics.

\begin{itemize}

\item \textbf{RQ1 – Semantic Extraction Quality:}  
To what extent does ontology-constrained prompting improve the semantic
correctness of extracted \emph{ESGMKG entities} compared to unconstrained baseline
LLM extraction?  
This question is measured using \emph{Semantic Accuracy}, which quantifies the
proportion of entities whose assigned ESGMKG types are semantically correct.

\item \textbf{RQ2 – Structural Validity and Ontology Compliance:}  
To what extent does the constructed knowledge graph conform to the structural
constraints defined by the ESGMKG ontology?  
This question is measured using \emph{Schema Compliance}, which captures
conformity to ontology rules (VR001--VR006) across entities, properties, and
relationships.

\item \textbf{RQ3 – Cost Efficiency:}  
What is the economic cost of producing validated ESG metric knowledge graphs,
and how much extraction effort is wasted on entities that fail validation?  
This question is measured using \emph{Cost per Entity} and \emph{Cost Waste
Ratio}, which together quantify API cost efficiency and extraction waste.

\end{itemize}

\subsection{Datasets and Experimental Setup}

We evaluate OntoMetric on five ESG regulatory documents widely used in practice
and exhibiting substantial variation in length, structure, and disclosure
style: two SASB industry standards (Commercial Banks and Semiconductors), one
international climate disclosure standard (IFRS~S2), one national adaptation
(AASB~S2 for Australia), and one cross-sector framework (TCFD Final Report).
Together, these documents span heterogeneous ESG reporting formats, ranging from
metric-centric standards with structured tables and explicit codes (SASB) to
principle-based narrative guidance with loosely specified indicators (TCFD).

All PDFs were processed using Stage~1 structure-aware segmentation with
TOC-guided boundaries and page-level provenance tracking. Each segment
corresponds to a coherent conceptual unit, such as a category definition, a
metric specification, or a calculation rule. Stage~2 extraction was applied
uniformly across all segments, followed by three-pass consolidation
(identifier resolution, entity deduplication, and relationship deduplication).
Stage~3 applies two-phase validation, comprising LLM-based semantic verification
and six rule-based ESGMKG schema validators (VR001--VR006). Knowledge graphs were
constructed using a uniform Python-based pipeline.

Stage~2 extraction and Phase~1 semantic verification in Stage~3 were performed
using a fixed large language model configuration (Claude~Sonnet~4.5, temperature
$=0.1$, \texttt{max\_tokens}$=16000$), which was held constant across all runs to
ensure stable and reproducible extraction behaviour.

Table~\ref{tab:documents-summary} summarises the evaluated documents together with
their page counts and the number of segments produced by Stage~1 segmentation.

\begin{table}[t]
\centering
\small
\caption{ESG regulatory documents used in the evaluation.}
\label{tab:documents-summary}
\begin{tabular}{@{}lcc@{}}
\toprule
\textbf{Document} & \textbf{Pages} & \textbf{Segments} \\
\midrule
SASB Commercial Banks & 23 & 10 \\
SASB Semiconductors   & 27 & 13 \\
IFRS~S2               & 46 & 10 \\
AASB~S2 (Australia)   & 58 &  8 \\
TCFD Report           & 74 & 19 \\
\bottomrule
\end{tabular}
\end{table}

To evaluate the end-to-end impact of ontology-guided extraction and validation
governance, we compare two extraction configurations applied to the same five
documents under identical runtime conditions. The two configurations differ only
in the presence of ontology-constrained prompting, provenance tracking, and
two-phase validation.

\begin{itemize}

\item \textbf{Baseline LLM Extraction.}  
Uses a minimal prompt that requests JSON
output containing entities and relationships but provides no ontology schema,
entity type definitions, relationship constraints, or extraction workflow.
Provenance tracking and validation are disabled.

\item \textbf{OntoMetric (Ontology-Constrained).}  
Uses a nine-component prompt (Table~\ref{tab:prompt-components}) embedding the ESGMKG schema, deterministic
identifier conventions, a four-step extraction workflow, and few-shot examples.
This configuration includes provenance tracking and two-phase validation
(semantic verification plus six schema rules VR001--VR006).

\end{itemize}

Both configurations use the same underlying language model and identical runtime
parameters to control for model capacity and stochasticity. All observed
differences in semantic correctness, structural validity, and cost efficiency
are attributable to the introduction of ontology-guided extraction and
validation governance as an integrated architectural intervention, rather than
to differences in model choice or parameter variation.

\subsection{Results and Discussion}
\label{subsec:results}

This section reports OntoMetric’s performance across five ESG regulatory
documents and contrasts ontology-guided extraction with baseline unconstrained
LLM extraction. Results are summarised in Table~\ref{tab:extraction-results} and
analysed along three independent dimensions corresponding to the three research
questions: semantic extraction quality (RQ1), structural validity under ontology
constraints (RQ2), and cost efficiency (RQ3).

\begin{table}[t]
\centering
\small
\caption{Extraction and validation results comparing ontology-guided and
baseline approaches. E~=~entities, R~=~relationships. ``---'' indicates that no
entities survived validation.}
\label{tab:extraction-results}
\resizebox{\columnwidth}{!}{%
\begin{tabular}{l|cc|cc}
\toprule
\textbf{Document} &
\multicolumn{2}{c|}{\textbf{Stage~2 (Extracted)}} &
\multicolumn{2}{c}{\textbf{Stage~3 (Validated)}} \\
\cmidrule(lr){2-3} \cmidrule(lr){4-5}
& \textbf{Ontology} & \textbf{Baseline} & \textbf{Ontology} & \textbf{Baseline} \\
\midrule
SASB Commercial Banks & 53E, 53R & 123E, 161R & 42E, 42R & 3E, 0R \\
SASB Semiconductors   & 69E, 71R & 149E, 200R & 62E, 64R & --- \\
IFRS~S2               & 80E, 65R & 126E, 176R & 68E, 58R & --- \\
AASB~S2 (Australia)   & 74E, 75R & 96E, 117R  & 66E, 65R & 3E, 0R \\
TCFD Report           & 88E, 86R & 235E, 286R & 57E, 54R & --- \\
\bottomrule
\end{tabular}}
\end{table}

\subsubsection{RQ1: Semantic Extraction Quality}

\begin{figure}[t]
  \centering
  \includegraphics[width=\linewidth]{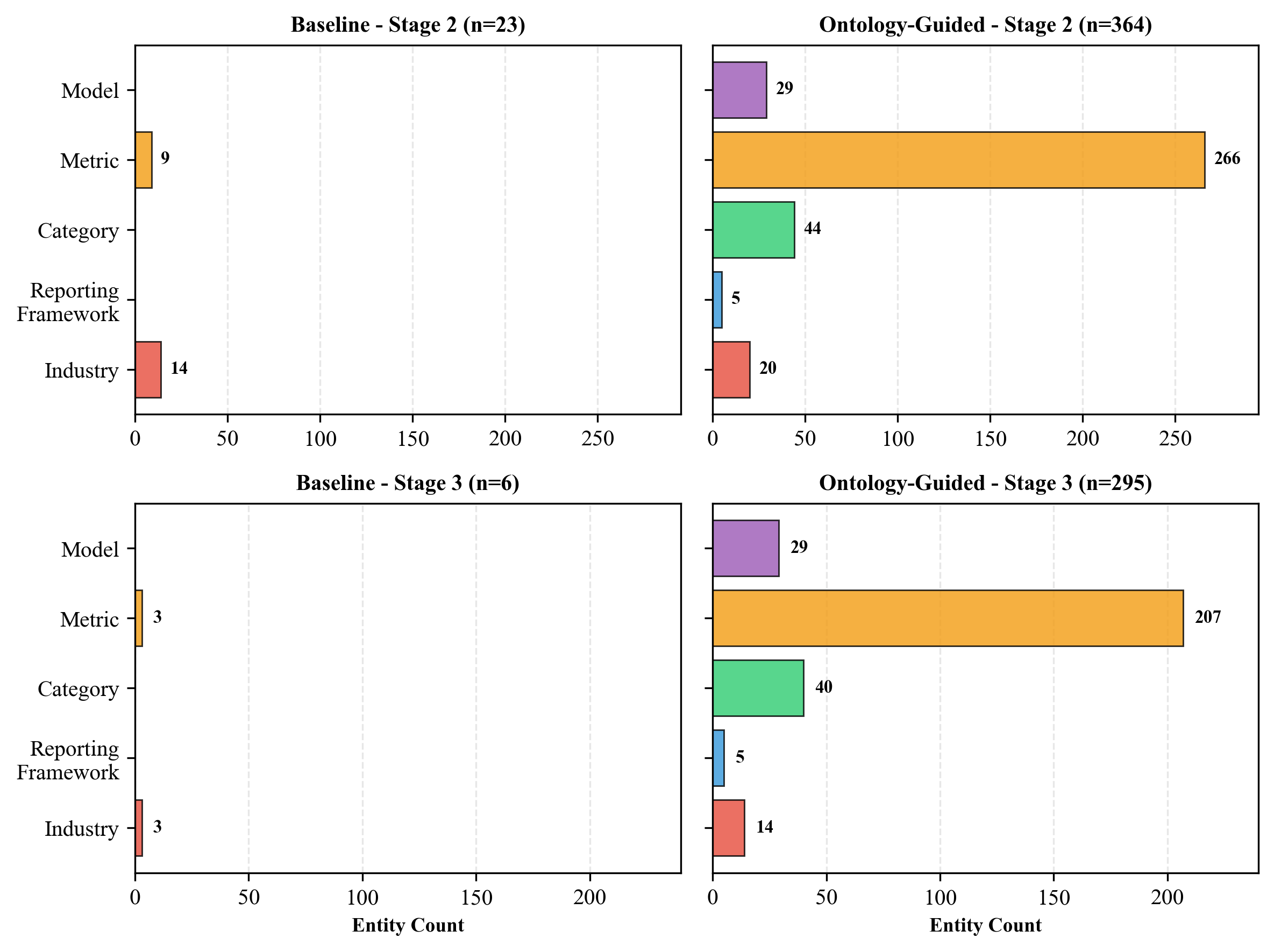}
  \caption{Entity type distribution comparison. Baseline extraction (left,
  $n=23 \rightarrow 6$) captures only two ESGMKG entity types, whereas
  ontology-guided extraction (right, $n=364 \rightarrow 295$) captures all five
  ESGMKG entity types across extraction (top) and validation (bottom) stages.}
  \label{fig:results-og}
\end{figure}

This subsection evaluates whether extracted entities are semantically correct
instances of ESGMKG entity types. Figure~\ref{fig:results-og} reports entity type
distributions across extraction and validation stages, and Figure~\ref{fig:validation-quality}
reports Semantic Accuracy values for ontology-guided extraction.

Ontology-guided extraction achieves Semantic Accuracy between 64.8\% and 89.9\%
across the five documents. The highest values are observed for SASB
Semiconductors (89.9\%) and AASB~S2 (89.2\%), while the lowest value occurs for
TCFD (64.8\%). These results indicate that most extracted entities are assigned
to semantically correct ESGMKG types and that their labels and descriptions are
consistent with the intended type semantics.

Figure~\ref{fig:results-og} shows that ontology-guided extraction identifies all
five ESGMKG entity types (Industry, ReportingFramework, Category, Metric, and
Model) across all documents. The distribution of entity types remains stable
between extraction and validation, with Metrics comprising the majority of
entities. This stability indicates that ontology-guided prompting constrains LLM
outputs toward semantically valid entity types at extraction time.

Baseline extraction exhibits extremely poor semantic quality. As shown in
Figure~\ref{fig:results-og}, baseline outputs capture only two of the five
ESGMKG entity types and produce large numbers of entities belonging to invalid
or unsupported types (e.g., \emph{Standard}, \emph{Organization}, \emph{Sector}).
These incorrect type assignments account for the near-zero Semantic Accuracy of
baseline extraction across all documents.

\subsubsection{RQ2: Structural Validity and Ontology Compliance}

\begin{figure*}[t]
  \centering
  \includegraphics[width=\linewidth]{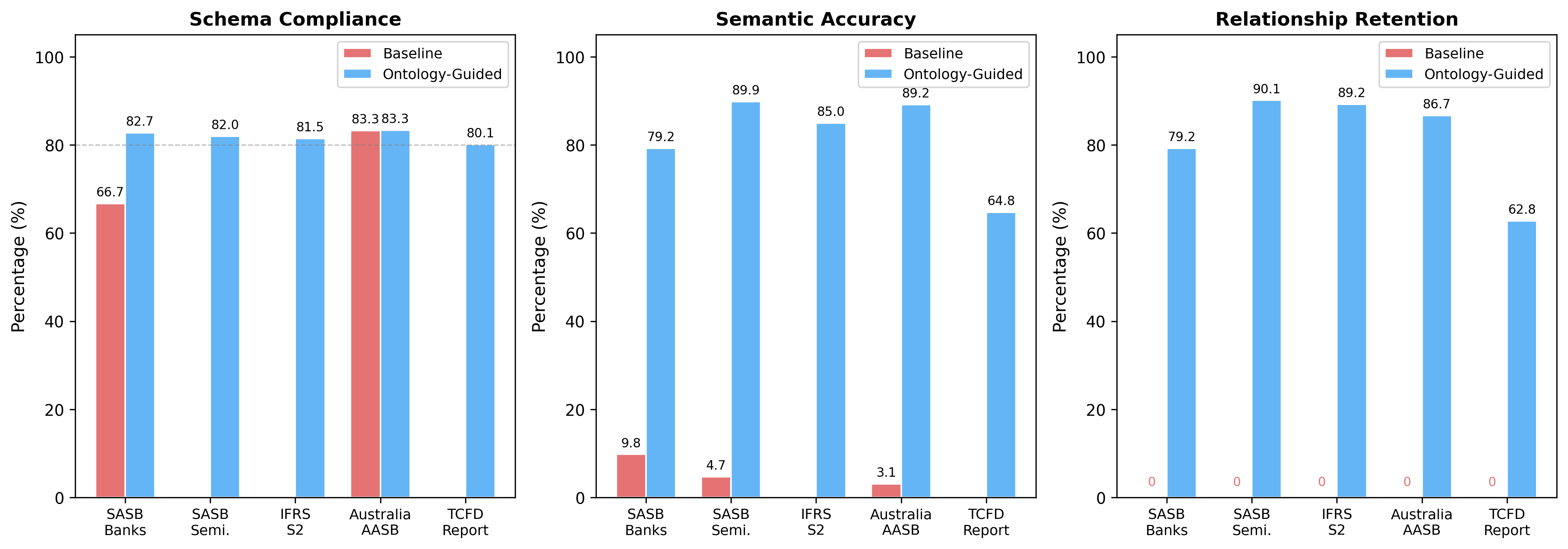}
  \caption{Structural validation metrics for ontology-guided extraction. Schema
  Compliance (80–83\%) and Relationship Retention (63–90\%) across five ESG
  documents. Semantic Accuracy is shown for completeness but is not used in RQ2
  analysis.}
  \label{fig:validation-quality}
\end{figure*}

This subsection evaluates whether extracted entities and relationships satisfy
ESGMKG structural constraints. Figure~\ref{fig:validation-quality} reports Schema
Compliance and Relationship Retention values for ontology-guided extraction.

Ontology-guided extraction achieves Schema Compliance between 80.1\% and 83.3\%
across all five documents, indicating that most extracted entities and
relationships satisfy ESGMKG schema rules, including required fields (VR002),
predicate validity (VR005), and cardinality constraints (VR006).

Relationship Retention ranges from 62.8\% to 90.1\%. Higher retention values are
observed for structured standards such as SASB Semiconductors (90.1\%) and
AASB~S2 (86.7\%), while lower retention occurs for narrative-heavy documents such
as TCFD (62.8\%). These results indicate that most extracted relationships are
structurally valid at extraction time and require only limited downstream
filtering.

Baseline extraction fails to produce structurally valid graphs. As shown in
Table~\ref{tab:extraction-results}, baseline outputs retain only 0–3.1\% of
entities after schema validation. The dominant failure modes are violations of
VR002 (missing required fields) and VR005 (invalid relationship predicates),
demonstrating that unconstrained LLM extraction does not satisfy ESGMKG
structural constraints.

\subsubsection{RQ3: Cost Efficiency}

\begin{figure}[t]
  \centering
  \includegraphics[width=\linewidth]{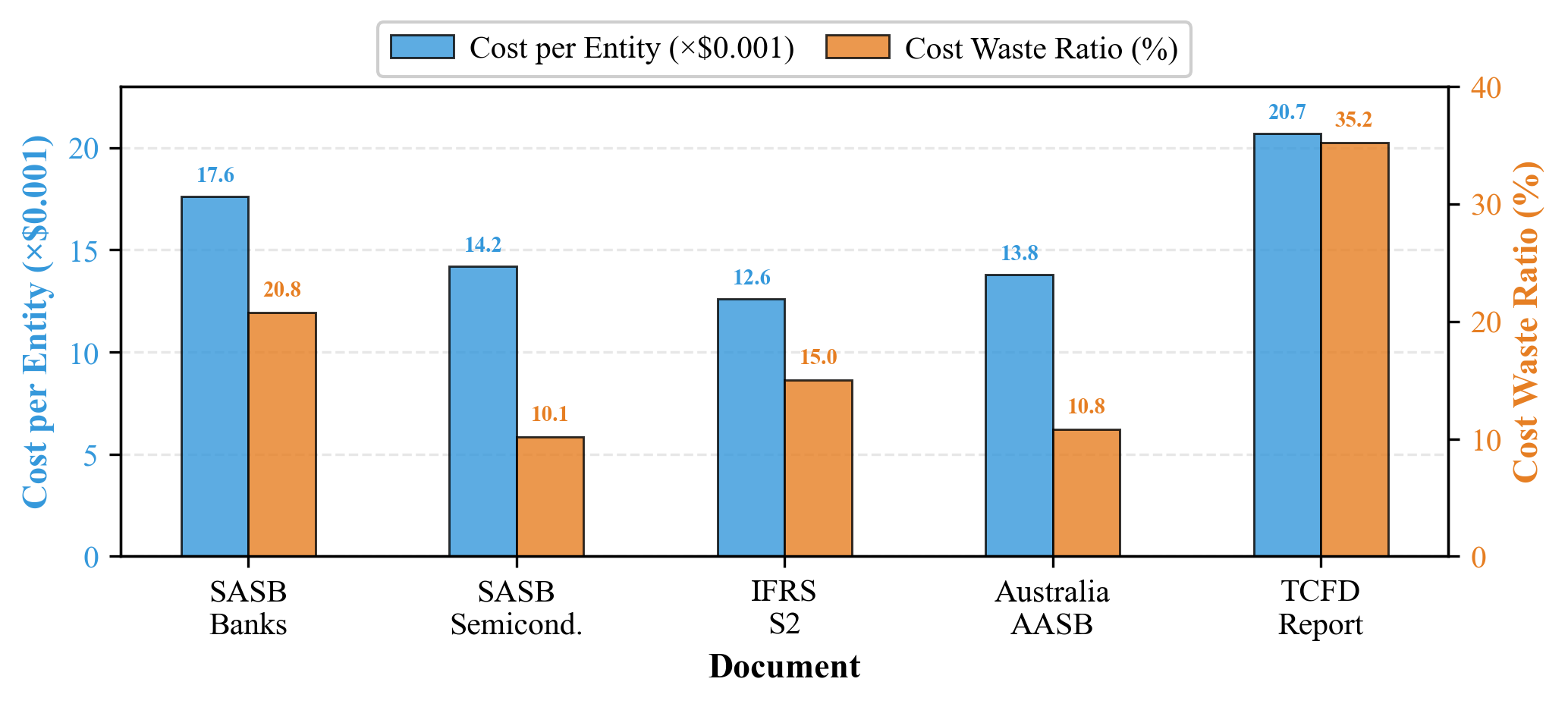}
  \caption{Per-document cost efficiency for ontology-guided extraction. Cost per
  validated entity ranges from \$0.013 to \$0.021, and Cost Waste Ratio ranges
  from 10\% to 35\% across five documents.}
  \label{fig:cost-per-document}
\end{figure}

\begin{figure}[t]
  \centering
  \includegraphics[width=\linewidth]{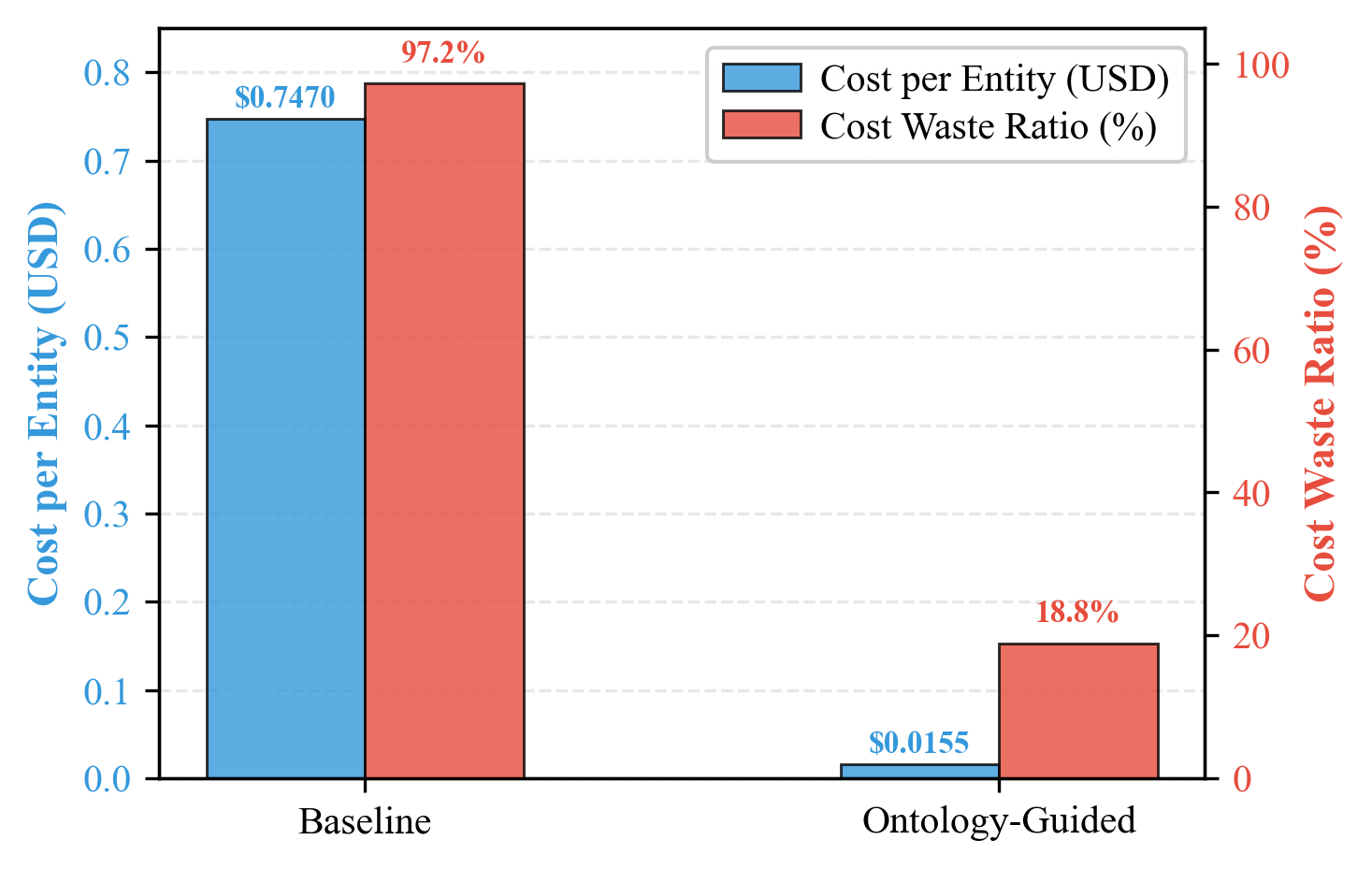}
  \caption{Method comparison. Baseline extraction yields six validated entities
  at \$0.747 per entity with 97\% cost waste, whereas ontology-guided extraction
  yields 295 validated entities at \$0.0155 per entity with 19\% waste, a
  48$\times$ efficiency improvement.}
  \label{fig:cost-method-comparison}
\end{figure}

This subsection evaluates the economic efficiency of ESG knowledge graph
construction under ontology-guided and baseline extraction. Figure~\ref{fig:cost-per-document}
reports per-document cost metrics for ontology-guided extraction, and
Figure~\ref{fig:cost-method-comparison} reports aggregate cost metrics for both
methods.

Ontology-guided extraction achieves stable cost efficiency across all five
documents. Cost per Validated Entity ranges from \$0.013 (IFRS~S2) to \$0.021
(TCFD). Cost Waste Ratio ranges from 10.1\% (SASB Semiconductors) to 35.2\%
(TCFD), reflecting the proportion of Stage~2 extraction effort spent on entities
subsequently filtered during validation.

Figure~\ref{fig:cost-method-comparison} shows that baseline extraction is
economically inefficient. Baseline extraction incurs a Cost Waste Ratio of
97.2\%, indicating that almost all API expenditure is spent on entities that do
not survive validation. In contrast, ontology-guided extraction incurs a Cost
Waste Ratio of 18.8\% and achieves a Cost per Validated Entity of \$0.0155,
representing a 48$\times$ improvement in cost efficiency.

\section{Conclusion and Future Work}

This paper addressed the automation and governance gap in the construction of
ESG metric knowledge graphs from unstructured regulatory documents. Although ESG
metric knowledge is inherently structured and graph-like—connecting industries,
reporting contexts, metric categories, metrics, calculation models, and input
dependencies—it does not exist in practice as an explicit, governed, or
machine-actionable artefact. Instead, this knowledge remains embedded implicitly
in regulatory standards and reporting frameworks. To close this gap, we
presented OntoMetric, an ontology-guided framework that operationalises the ESG
Metric Knowledge Graph (ESGMKG) ontology as a first-class constraint for the
automated construction and governance of ESG metric knowledge graphs.

OntoMetric treats ESGMKG not merely as a schema for post-hoc validation, but as a
semantic backbone embedded directly into the extraction and population process.
By integrating structure-aware segmentation, ontology-constrained LLM
extraction enriched with semantic fields, and two-phase validation combining
semantic type verification with rule-based schema checking, OntoMetric enables
the scalable population of ESGMKG with explicit representations of ESG metric
definitions, scope conditions, units, calculation models, and compositional
dependencies. Throughout this process, segment-level and page-level provenance
is preserved, ensuring that every metric, model, and dependency remains
traceable to authoritative regulatory text.

Empirical evaluation on five major ESG regulatory documents demonstrated that
ontology-guided extraction substantially improves the reliability of ESG metric
knowledge graph construction compared with unconstrained LLM extraction.
OntoMetric achieved 65--90\% semantic accuracy and over 80\% schema compliance
across all evaluated documents, enabling the automated instantiation of
semantically faithful and structurally valid ESG metric knowledge graphs with
explicit audit trails.

Three design principles were central to this outcome. First, enriching ESGMKG
entities with semantic fields such as labels, descriptions, and domain
properties supports machine-actionable representations of ESG metric knowledge,
rather than treating metrics as opaque identifiers. Second, separating
validation into semantic type verification and structural rule enforcement
isolates two distinct failure modes—hallucinated ESG concepts versus ontology
violations—providing systematic governance over populated graph instances.
Third, preserving provenance during extraction ensures traceability and
auditability, which are essential for regulatory and compliance contexts.

Several limitations motivate future work. OntoMetric currently relies on
table-of-contents structures for segmentation; extending this mechanism to
documents with implicit, inconsistent, or weakly signalled section boundaries
is a natural next step. The framework is evaluated on five English-language ESG
regulatory standards and does not yet cover corporate reports, financial
filings, or non-English sources, which may exhibit different structural and
linguistic characteristics. OntoMetric operationalises five ESGMKG entity types
(Industry, ReportingFramework, Category, Metric, Model), omitting other ESG
concepts such as risks, opportunities, targets, governance structures, and
temporal aspects that are present in many standards and reports. Semantic
validation currently relies on the same class of large language model used for
extraction, introducing potential circularity in which systematic model biases
may go undetected, while schema validation enforces structural conformance but
does not assess factual correctness. The pipeline also processes only textual
content and does not incorporate information from charts, figures, or embedded
tables beyond basic multi-page merging. Although ontology-guided extraction
substantially reduces wasted computation relative to unconstrained LLM
extraction, OntoMetric still depends on multiple LLM calls for extraction and
semantic validation, introducing nontrivial operational costs for large-scale
or frequently updated corpora, and currently lacks incremental processing and
caching mechanisms for efficient knowledge graph maintenance.

Overall, this work demonstrates that ESG metric knowledge graphs can be
constructed automatically from regulatory documents when ontological structure
and compositional constraints are treated as first-class elements of the
population process. OntoMetric provides a principled and practical foundation
for the scalable, auditable, and ontology-driven construction of ESG metric
knowledge graphs from authoritative regulatory sources.

\bibliographystyle{ACM-Reference-Format}
\bibliography{main}

\clearpage
\appendix

\section{Algorithm Specifications}

This appendix summarises the three-stage OntoMetric pipeline in algorithmic
form. The descriptions are consistent with the implementation described in Section~3.

\subsection{Stage~1 -- Structure-aware Segmentation}
\label{app: stage-1-algorithm}
Stage~1 converts each regulatory PDF into a set of semantically coherent
segments, preserving page-level provenance and table structure. The output of
this stage is a JSON file that serves as input to ontology-guided extraction.

\begin{table}[ht]
\centering
\small
\caption{Algorithm 1: Segmentation (Stage~1).}
\label{alg:stage1-segmentation}
\begin{tabular}{@{}p{\columnwidth}@{}}
\toprule
\textbf{Input:} PDF document $D$ \\
\textbf{Output:} Segments $S = \{s_1, \ldots, s_n\}$ \\
\midrule
1.\quad $title \gets \textsc{ExtractTitle}(D,\ \text{page}=1)$ \\[2pt]
2.\quad $tocPage \gets \textsc{FindTOCPage}(D)$ \\[2pt]
3.\quad $toc \gets \textsc{ParseTOC}(D,\ \text{page}=tocPage)$ \\[2pt]
4.\quad \textbf{for each} section $sec$ in $toc$ \textbf{do} \\[2pt]
\hspace*{1.2em}4.1\quad $s.id \gets \textsc{GenerateID}(sec.number)$ \\[-1pt]
\hspace*{1.2em}4.2\quad $s.title \gets sec.title$ \\[-1pt]
\hspace*{1.2em}4.3\quad $s.pages \gets [sec.start, sec.end]$ \\[-1pt]
\hspace*{1.2em}4.4\quad $s.content \gets \textsc{ExtractText}(D,\ s.pages)$ \\[-1pt]
\hspace*{1.2em}4.5\quad $s.tables \gets \textsc{ExtractTables}(D,\ s.pages)$ \\[-1pt]
\hspace*{1.2em}4.6\quad $s.tables \gets \textsc{MergeMultiPageTables}(s.tables)$ \quad \textit{// merge split tables} \\[-1pt]
\hspace*{1.2em}4.7\quad $s.content \gets \textsc{CleanText}(s.content)$ \quad \textit{// remove headers/footers, noise} \\[-1pt]
\hspace*{1.2em}4.8\quad $S \gets S \cup \{s\}$ \\[2pt]
5.\quad \textbf{return} $S$ \\
\bottomrule
\end{tabular}
\end{table}

\subsection{Stage~2 -- Ontology-Guided LLM Extraction}
\label{app: stage2-Algorithm}
Stage~2 takes the segmented JSON and applies ontology-guided prompts to
Claude~4.5~Sonnet. It parses the JSON response into entities and relationships,
attaches provenance, and aggregates outputs across segments.

\begin{table}[ht]
\centering
\small
\caption{Algorithm 2: Ontology-Guided Extraction (Stage~2).}
\label{alg:stage2}
\begin{tabular}{@{}p{\columnwidth}@{}}
\toprule
\textbf{Input:} Segments $S$, ontology $\mathcal{O}$, LLM $\mathcal{L}$ \\
\textbf{Output:} Raw extracted entities $\mathcal{E}$ and relationships $\mathcal{R}$ \\
\midrule
1.\quad Initialise sets $\mathcal{E} \gets \emptyset$, $\mathcal{R} \gets \emptyset$. \\[3pt]

2.\quad \textbf{for each} segment $s \in S$ \textbf{do} \\[1pt]
\hspace*{1.2em}2.1\quad $prompt \gets BuildPrompt(s, \mathcal{O})$ \\[-1pt]
\hspace*{1.2em}2.2\quad $resp \gets \mathcal{L}(prompt,\ \text{temperature}=0.1)$ \\[-1pt]
\hspace*{1.2em}2.3\quad $ext \gets ParseJSON(resp)$ \\[-1pt]

\hspace*{1.2em}2.4\quad \textbf{for each} entity $e \in ext.entities$ \textbf{do} \\[-1pt]
\hspace*{2.4em}$e.prov \gets \{s.id,\ s.title,\ s.pages\}$ \quad \textit{// attach provenance} \\[-1pt]
\hspace*{2.4em}$\mathcal{E} \gets \mathcal{E} \cup \{e\}$ \\[-1pt]

\hspace*{1.2em}2.5\quad \textbf{for each} relationship $r \in ext.relationships$ \textbf{do} \\[-1pt]
\hspace*{2.4em}$r.prov \gets \{s.id,\ s.title,\ s.pages\}$ \\[-1pt]
\hspace*{2.4em}$\mathcal{R} \gets \mathcal{R} \cup \{r\}$ \\[4pt]

3.\quad \textbf{return} $(\mathcal{E}, \mathcal{R})$ \\
\bottomrule
\end{tabular}
\end{table}

\subsection{Stage~3 -- Two-Phase Validation}
\label{app: stage3-Algorithm}
Stage~3 performs two-phase validation: (1)~semantic validation using an LLM gate-keeper to verify entity type correctness, and (2)~schema validation enforcing six ontology compliance rules (VR001--VR006). It computes the evaluation metrics used in Section~4 (Schema Compliance, Semantic Accuracy, Relationship Retention), and produces a validated knowledge graph plus a detailed validation report.

\begin{table}[ht]
\centering
\small
\caption{Algorithm 3: Two-Phase Knowledge Graph Validation (Stage~3).}
\label{alg:stage3-validation}
\begin{tabular}{@{}p{\columnwidth}@{}}
\toprule
\textbf{Input:} Graph $\mathcal{G} = (\mathcal{E}, \mathcal{R})$, 
rules $\mathcal{V} = \{v_1,\ldots,v_6\}$, 
LLM $\mathcal{L}$ \\
\textbf{Output:} Validated graph $\mathcal{G}' = (\mathcal{E}', \mathcal{R}')$, metrics $M$ \\
\midrule

\multicolumn{1}{@{}l@{}}{\textbf{\textit{Phase 1: Semantic Validation (LLM gate)}}} \\[3pt]

1.\quad \textbf{for each} entity $e$ in $\mathcal{E}$ \textbf{do} \\[1pt]
\hspace*{1.2em}1.1\quad $isCorrect \gets \mathcal{L}(\text{ValidateSemanticType}(e))$ \\[-1pt]
\hspace*{1.2em}1.2\quad \textbf{if} $\neg isCorrect$ \textbf{then} mark $e$ for removal \\[1pt]

2.\quad $\mathcal{E} \gets \mathcal{E} \setminus \{\text{marked entities}\}$ \\[1pt]

3.\quad $\mathcal{R} \gets \{ r \in \mathcal{R} : r.subject, r.object \in \mathcal{E} \}$ 
\quad \textit{// cascade filtering} \\[6pt]

\multicolumn{1}{@{}l@{}}{\textbf{\textit{Phase 2: Schema Validation (six rules)}}} \\[3pt]

4.\quad \textbf{initialise}\quad $violations \gets []$ \\[1pt]

5.\quad \textbf{for each} rule $v_i$ in $\mathcal{V}$ \textbf{do} \\[1pt]
\hspace*{1.2em}5.1\quad $viol_i \gets ExecuteRule(v_i, \mathcal{E}, \mathcal{R})$ \\[-1pt]
\hspace*{1.2em}5.2\quad \textbf{if} $|viol_i| > 0$ \textbf{then} 
$violations \gets violations + viol_i$ \\[6pt]

6.\quad $\mathcal{E}' \gets \{ e \in \mathcal{E} : e.id \notin violations \}$ \\[1pt]

7.\quad $\mathcal{R}' \gets 
\{ r \in \mathcal{R} : r.subject, r.object \in \mathcal{E}' \}$ \\[6pt]

\multicolumn{1}{@{}l@{}}{\textbf{\textit{Quality metrics computation}}} \\[3pt]

8.\quad $M.\text{schemaCompliance} 
\gets \frac{1}{6} \sum_{i=1}^{6} \frac{\text{items passing } v_i}{\text{total items for } v_i} \times 100$ \\[1pt]

9.\quad $M.\text{semanticAccuracy} 
\gets \frac{|\mathcal{E}|}{|\mathcal{E}_{\text{original}}|} \times 100$
\quad \textit{// from Phase 1} \\[1pt]

10.\quad $M.\text{relationshipRetention} 
\gets \frac{|\mathcal{R}'|}{|\mathcal{R}_{\text{original}}|} \times 100$ \\[6pt]

11.\quad \textbf{return} $\mathcal{G}' = (\mathcal{E}', \mathcal{R}')$, $M$ \\
\bottomrule
\end{tabular}
\end{table}

\clearpage
\section{Appendix: Validated ESG Metric Knowledge Graph Instances}

This appendix presents five validated ESG Metric Knowledge Graph (ESGMKG)
instances generated by OntoMetric from real-world regulatory standards:
SASB Commercial Banks, SASB Semiconductors, IFRS~S2, AASB~S2 (Australia), and the
TCFD Final Report.

Each figure visualises the final knowledge graph after two-phase validation,
showing ESGMKG entity types (Industry, ReportingFramework, Category, Metric,
Model) and their relationships, including compositional dependencies between
metrics, calculation models, and input variables. These figures provide
qualitative evidence of how heterogeneous regulatory documents are transformed
into a unified, ontology-governed representation.

\subsection{SASB Commercial Banks}
\label{app:kg-sasb-banks}
\vspace{\baselineskip}

\begin{figure}[!htbp]
  \includegraphics[width=\linewidth]{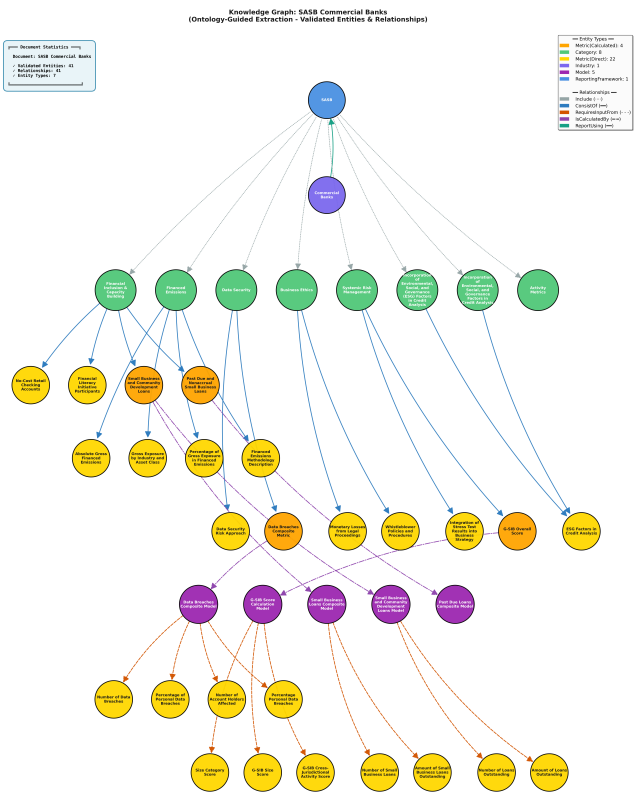}
  \captionof{figure}{Validated ESG metric knowledge graph instance for \emph{SASB Commercial Banks} produced by OntoMetric.}
  \label{fig:app-kg-sasb-banks}
\end{figure}


\subsection{SASB Semiconductors}
\label{app:kg-sasb-semi}
\vspace{\baselineskip}

\begin{center}
  \includegraphics[width=\linewidth]{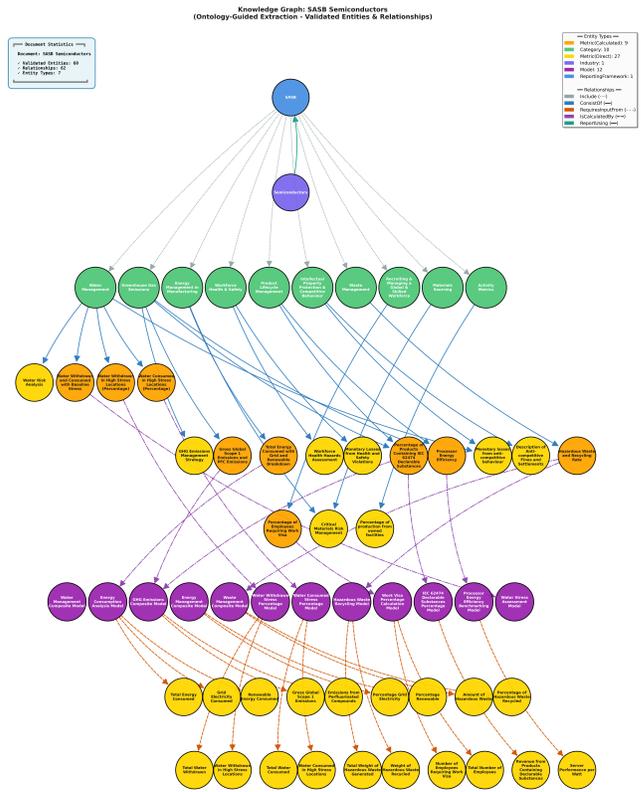}
  \captionof{figure}{Validated ESG metric knowledge graph instance for \emph{SASB Semiconductors} produced by OntoMetric.}
  \label{fig:app-kg-sasb-semi}
\end{center}


\subsection{IFRS~S2}
\label{app:kg-ifrs-s2}
\vspace{\baselineskip}

\begin{center}
  \includegraphics[width=\linewidth]{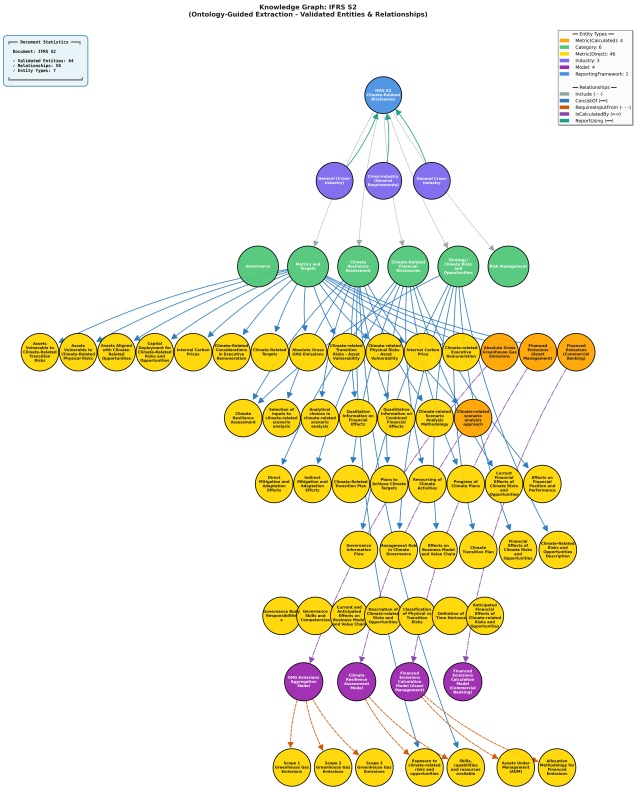}
  \captionof{figure}{Validated ESG metric knowledge graph instance for \emph{IFRS~S2} produced by OntoMetric.}
  \label{fig:app-kg-ifrs-s2}
\end{center}


\subsection{AASB~S2 (Australia)}
\label{app:kg-aasb-s2}
\vspace{\baselineskip}

\begin{center}
  \includegraphics[width=\linewidth]{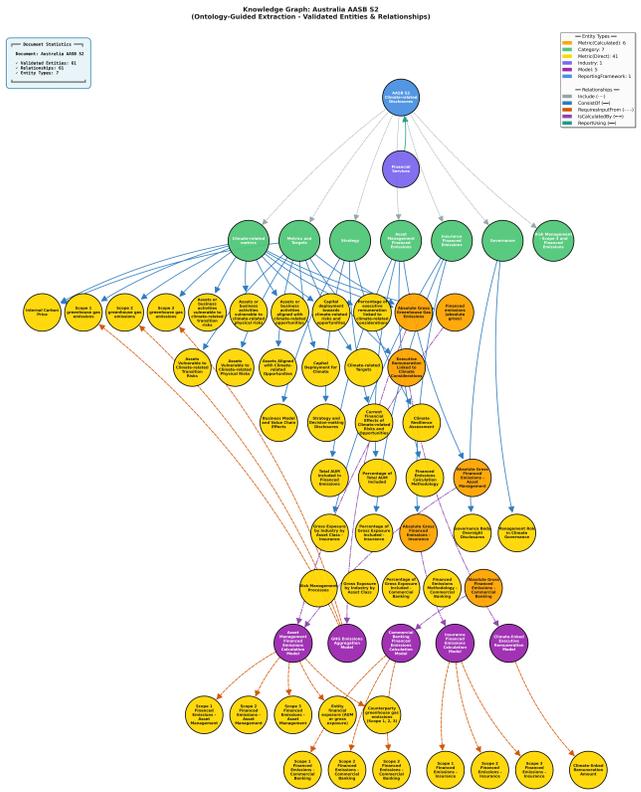}
  \captionof{figure}{Validated ESG metric knowledge graph instance for \emph{AASB~S2 (Australia)} produced by OntoMetric.}
  \label{fig:app-kg-aasb-s2}
\end{center}

\clearpage

\twocolumn[{
\subsection{TCFD Final Report}
\label{app:kg-tcfd}

\centering
\includegraphics[width=0.9\textwidth,keepaspectratio]{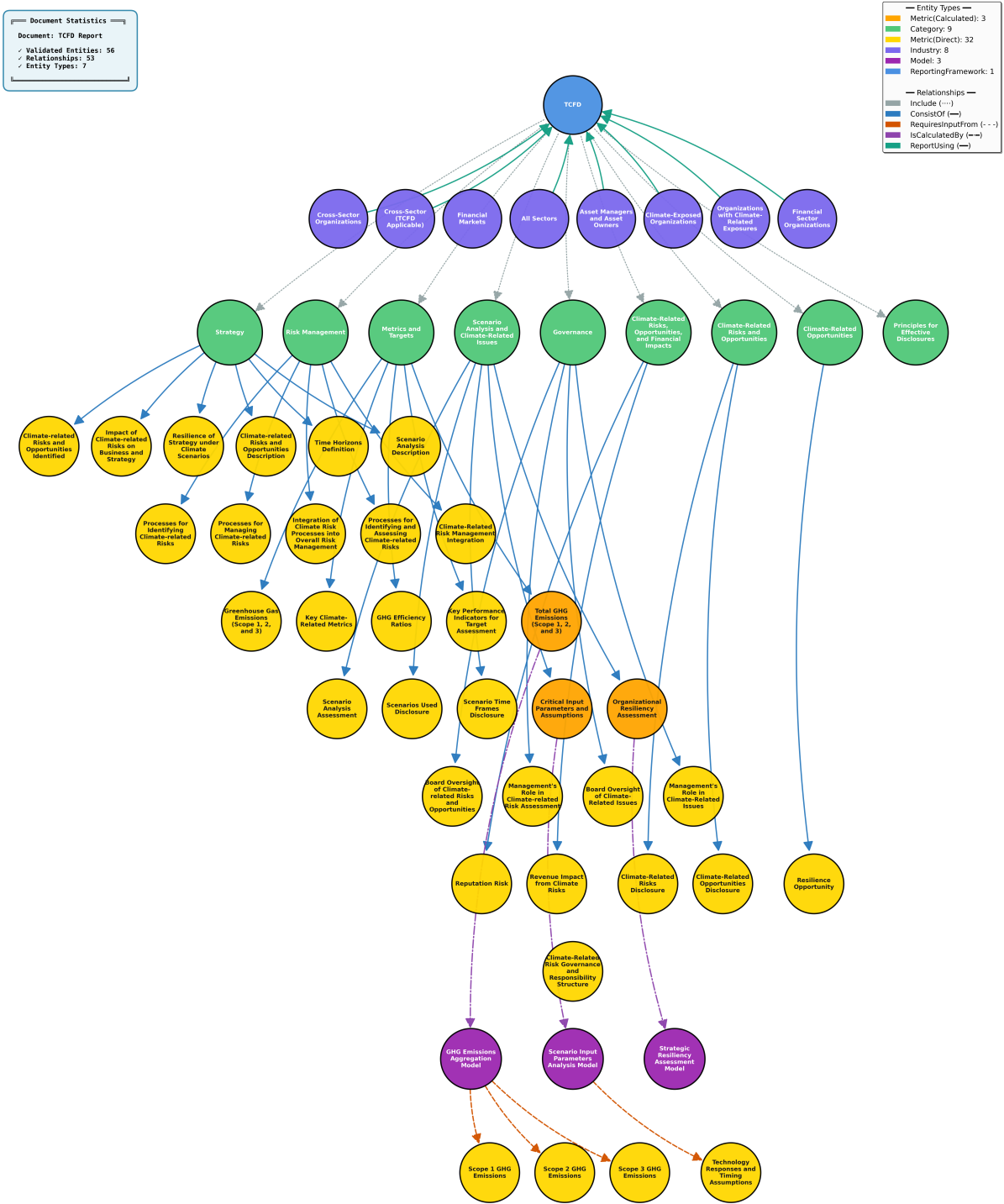}
\captionof{figure}{Validated ESG metric knowledge graph instance for \emph{TCFD Final Report} produced by OntoMetric.}
\label{fig:app-kg-tcfd}
\vspace{1em}
}]

\end{document}